\documentclass[letterpaper, 10 pt, conference]{ieeeconf}  
\usepackage{graphicx}
\IEEEoverridecommandlockouts                              

\usepackage{amsmath}
\usepackage{amssymb}
\usepackage{graphics}
\usepackage[hidelinks]{hyperref}
\usepackage{xcolor}
\usepackage{dblfloatfix}
\usepackage {tikz}
\usepackage{mathtools}
\usepackage{svg}
\usepackage[bottom]{footmisc}
\usepackage{amsmath}
\usepackage{soul}
\usepackage[switch]{lineno}
\usepackage{multirow}
\usepackage{caption}
\usetikzlibrary {positioning}

\title{\LARGE \bf
Influence of Team Interactions on Multi-Robot Cooperation: \\ A Relational Network Perspective
}

\author{Yasin Findik$^{1*}$, Hamid Osooli$^{1*}$, Paul Robinette$^{2}$, Kshitij Jerath$^{3}$, and  S. Reza Ahmadzadeh$^{1}$
\thanks{$^{*}$indicates equal contribution}
\thanks{$^{1}$ PeARL Lab, Richard Miner School of Computer and Information Sciences, University of Massachusetts Lowell, MA, USA {\tt\small \{yasin\_findik, hamid\_osooli\}@student.uml.edu, reza@cs.uml.edu}}%
\thanks{$^{2}$ Department of Electrical and Computer Engineering, University of Massachusetts Lowell, MA, USA {\tt\small paul\_robinette@uml.edu}}%
\thanks{$^{3}$ Department of Mechanical Engineering, University of Massachusetts Lowell, MA, USA {\tt\small kshitij\_jerath@uml.edu}}
}%
\begin{document}

\maketitle
\thispagestyle{empty}
\pagestyle{empty}
\begin{abstract}
Relational networks within a team play a critical role in the performance of many real-world multi-robot systems. To successfully accomplish tasks that require cooperation and coordination, different agents (e.g., robots) necessitate different priorities based on their positioning within the team. Yet, many of the existing multi-robot cooperation algorithms regard agents as interchangeable and lack a mechanism to guide the type of cooperation strategy the agents should exhibit. To account for the team structure in cooperative tasks, we propose a novel algorithm that uses a relational network comprising inter-agent relationships to prioritize certain agents over others. Through appropriate design of the team's relational network, we can guide the cooperation strategy, resulting in the emergence of new behaviors that accomplish the specified task. We conducted six experiments in a multi-robot setting with a cooperative task. Our results demonstrate that the proposed method can effectively influence the type of solution that the algorithm converges to by specifying the relationships between the agents, making it a promising approach for tasks that require cooperation among agents with a specified team structure.
\end{abstract}

\section{Introduction}


Effective coordination and cooperation among multiple agents (e.g., robots) is imperative to accomplishing collective or individual objectives~\cite{busoniu2008comprehensive} in a range of applications, including autonomous driving~\cite{bresson2017simultaneous, grigorescu2020survey}, search \& rescue~\cite{calisi2005autonomous, kleiner2006rfid}, and logistics \& transportation~\cite{diaz2021editorial}.  In recent years, deep learning techniques within multi-agent reinforcement learning (MARL) approaches have shown promising results in addressing the cooperation challenges in multi-agent environments~\cite{huttenrauch2019deep, levine2016end, mnih2015human}. One of the most prominent paradigms in MARL for cooperative tasks is the centralized training with decentralized execution (CTDE)~\cite{oliehoek2008optimal}.

State-of-the-art CTDE approaches developed in recent years are widely recognized for tackling coordination problems~\cite{sunehag2017value, rashid2020monotonic, son2019qtran, lowe2017multi, yu2021surprising}. These CTDE approaches offered no control over which coordination strategy the algorithms yield, which is because of the absence of a mechanism for specifying the particular type of agent-agent relations. To address this issue, some algorithms borrowed and applied the idea of coordination graph~\cite{guestrin2001multiagent} to share action values among the agents using deep learning~\cite{bohmer2020deep}. This approach requires the algorithm to learn how to weigh and share action values from an undirected graph, which does not allow for the influence of certain behaviors among others.
We posit that one way to design cooperation strategies that account for team structure is to define a relational network represented with a directed graph, which captures the importance and priorities agents put on each other. We propose a novel CTDE-based algorithm that exerts influence on the type of cooperation strategy by utilizing a relational network that determines the relationships in the team. Our proposed algorithm utilizes this relational network to prioritize certain agents over others,
enabling agents to uncover new behaviors that successfully accomplish the task while maintaining the team structure.

\begin{figure}[t]
\centering
    \includegraphics[trim={0 0em 0 0},clip, width=0.9\columnwidth]{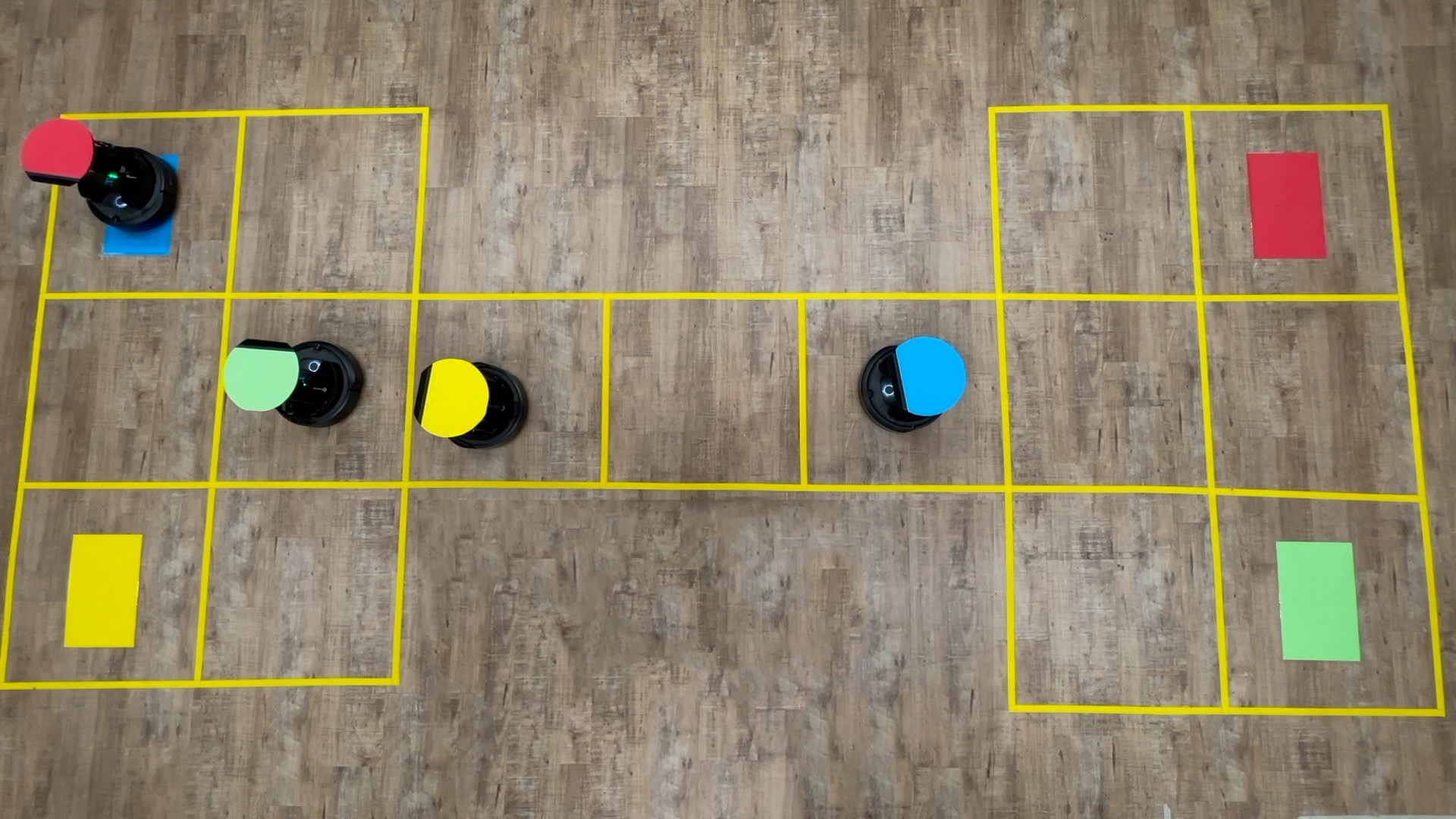}
  \caption{\small{Our experimental setup where four Turtlebot4 robots learn to navigate the \textit{Switch} environment cooperatively so each one can reach the goal location matching their color.}}
  \label{fig:robots_in_action}
\end{figure}




We conducted six experiments in both simulation and real-world settings, using our algorithm in a multi-robot task in which coordination among agents was influenced through different relational networks. We demonstrated that our proposed method can effectively steer the algorithm to converge to a specific team behavior governed by the relational network among agents. Unlike most methods, the proposed approach enables effective cooperation in a multi-agent team and allows for the emergence of different behaviors by leveraging relational networks. 

\section{Related Work}

Multi-agent reinforcement learning (MARL) in cooperative settings has become an active area of research in recent years, with various approaches being explored for successfully training agents to cooperate towards a shared goal. One such approach is fully decentralized learning, where each agent learns its own policy independently. In these approaches, the cooperative behavior emerges as the learned policies are employed in the environment. For example, Independent Q-learning~\cite{tan1993multi} learns a separate action-value table for each agent via Q-learning~\cite{watkins1992q}. Since tabular Q-learning is limited in tackling high-dimensional state and action space, the Independent Q-learning framework was later extended with function approximation~\cite{tampuu2017multiagent}. However, independent learning in multi-agent settings suffers from the non-stationarity of the problem caused by other agents' actions from the perspective of one agent. 
Because the Markov property is no longer valid in non-stationary environments, the convergence of Q-learning based decentralized algorithms cannot be guaranteed~\cite{hernandez2017survey}. Later, the introduction of fully decentralized algorithms with networked agents~\cite{zhang2018fully}, which could communicate and share their observations and actions, resulted in a method that could tackle the non-stationarity problem. 
In frameworks with networked agents, an undirected graph governs the connections between agents. 
Similarly, we also leverage a graph to represent some form of rapport between the agents. However, unlike frameworks that focus on the use of a communication channel~\cite{zhang2018fully},  our graph symbolizes and establishes relations between agents, effectively creating a relational network that directly affects the team behavior. 

Another approach to MARL in cooperative settings is fully centralized learning, where all agents share a single controller and learn a joint policy or value function together~\cite{claus1998dynamics}. However, this approach is computationally expensive and can become intractable since the observation and action space grow exponentially with the number of agents.
To address the limitations of both fully centralized and fully decentralized learning in cooperative MARL settings, a novel paradigm known as centralized training with decentralized execution (CTDE)~\cite{oliehoek2008optimal} has been proposed. In this approach, individual agents carry out their actions while a centralized mechanism integrates their strategies. Both policy-based and value-based methods have been successfully applied in implementing the CTDE paradigm. During training, policy-based methods such as Multi-Agent Deep Deterministic Policy Gradient (MADDPG)~\cite{lowe2017multi} and Multi-Agent Proximal Policy Optimization (MAPPO)~\cite{yu2021surprising} allow the critic to consider global observations of all agents, while value-based techniques such as Value Decomposition Networks (VDN)~\cite{sunehag2017value}, QMIX~\cite{rashid2020monotonic}, and QTRAN~\cite{son2019qtran} improve upon Q-Learning ~\cite{watkins1992q} by incorporating a centralized function that calculates the joint Q-value from the individual action-values of each agent.

From fully decentralized paradigm to CTDE, previous approaches usually have focused on converging to an optimal solution for solving cooperation problems in MARL. However, the ability of a multi-robot team to arrive at an optimal or sub-optimal solution is strongly influenced by the underlying team interactions~\cite{findik2023impact, haeri2022reward}. Unlike~\cite{haeri2022reward}, our approach does not rely on reward sharing among agents to alter their individual rewards based on the relationship. Instead, our method entails modifying the way agents contribute to the team reward, which is determined by the relational network. Ultimately, agents  still receive the same individual reward provided by the environment.


To better understand these effects, we introduce the integration of the relational network technique, which effectively captures such team interactions and rewards, into the CTDE framework.
We choose to explore and study this idea with VDN, as it is a simple yet effective CTDE approach for learning cooperative behaviors. However, our proposed notion of relational networks can be applied to other CTDE approaches as well.

\section{Methodology}

\subsection{Markov Decision Process}
We characterized Decentralized Markov Decision Process as a tuple $\langle  \mathcal{S}, \mathcal{A}, \mathcal{R}, \mathcal{T}, n, \gamma\rangle$ where $s \in \mathcal{S}$ indicates the true state of the environment, the joint set of individual actions and rewards are represented by $\mathcal{A} \coloneqq \{a_1, a_2, \dots, a_n \}$, $\mathcal{R} \coloneqq \{r_1, r_2, \dots, r_n \}$, respectively,  $\mathcal{T} (s, A, s') \colon \mathcal{S} \times \mathcal{A}\times \mathcal{S} \mapsto [1,0]$ is the dynamics function defining the transition probability,  
$n$ is the the number of agents, and $\gamma\in[0,1)$ is the discount factor. 

\subsection{Value Decomposition Networks}

In this section, we provide a brief background on the Value-Decomposition Networks (VDN)~\cite{sunehag2017value} upon which we built our proposed method. VDN falls under the category of CTDE approaches. It addresses the non-stationarity problem of decentralized learning effectively through centralized training and it addresses the scalability issue of centralized learning through decentralized execution. 

VDN maintains a separate action-value function for each agent $i \in \{ 1,...,n\}$, denoted as $Q_i(s, a_i)=\mathbb{E}[G | S=s, A=a_i]$ and approximated as $\hat{Q}_i(s, a_i, w_i)$ where $w_i$ is the weight vector and $G$ is the return. VDN combines these individual $Q_i$ values to obtain the central action value $Q_{tot}$, as
\begin{align}
\label{q_tot}
Q_{tot} = \sum_{i=1}^{n} \hat{Q}_i(s, a_i, w_i).
\end{align}
The ultimate goal is to minimize the Temporal Difference (TD)~\cite{sutton1988learning} error:
\begin{align}
\label{td_error}
e_{\textrm{TD}} = r_{\textrm{team}} + \gamma \max_{u'}(Q_{\textrm{tot}}(s', u')) - Q_{\textrm{tot}}(s, u), 
\end{align}
\noindent where $u$ is the joint action of the agents and $r_{\textrm{team}}$ 
is the sum of agents' rewards with uniform weights. The algorithm minimizes the TD error by backpropagating $e_{\textrm{TD}}^2$ to each agent's model. This process allows the agents to coordinate their actions towards  maximizing the team reward, which is of utmost importance in most cooperative settings. 
The central aspect of CTDE paradigm becomes evident during training, where the agent networks are trained with a centralized $Q_{tot}$ that is computed from the sum of individual $Q_i$ values. Whereas, during execution, each agent's actions are determined by their own neural network, rendering the execution decentralized.

\subsection{Graph-based Value Decomposition Network}

Most MARL problems typically have multiple solutions with different levels of optimality. Although VDN and several other approaches do not provide a guarantee of finding the global optimal solution for every possible MARL environment or task, in general, they aim to maximize team rewards by improving coordination and cooperation among agents, and converge to one of several possible solutions (potentially the global optimal solution). However, if there are multiple (optimal) solutions, the randomness of agent's exploration determines the solution to which the algorithm converges. To be able to influence the team behavior and have more control over the dominant solution, we introduce Graph-based VDN (G-VDN). 
Similar to VDN, G-VDN utilizes an individual action-value function $Q_i$ approximator for each agent and a summation function that aggregates the distinct $Q_i$ values. The difference between the two, however, lies in the approach they use to aggregate team rewards. In VDN, every agent contributes their individual rewards to the team, uniformly. On the other hand, G-VDN aggregates individual rewards by considering the relational network between agents in computing joint rewards, 
thereby shaping the converged behavior by taking inter-agent relationships into account.


Crucially, we modify the MDP to additionally include the relational network which is represented by a directed graph $\mathcal{G}=(\mathcal{V}, \mathcal{E}, \mathcal{W})$, where each agent $i \in \{ 1,...,n\}$ is a vertex $v_i$, $\mathcal{E}$ is the set of directed edges $e_{ij}$ directed from $v_i$ to $v_j$, and the weight of the edges is represented by the matrix $\mathcal{W}$ with elements $\omega_{ij}\in[0, 1]$ for each edge. The direction of the edges and the associated weights signify the relationship between the agents, with an edge $e_{ij}$ directed from $i$ to $j$ indicating that agent $i$ cares about or is vested in the outcomes for agent $j$. Based on the modified MDP represented as a tuple $\langle  \mathcal{S}, \mathcal{A}, \mathcal{R}, \mathcal{T}, \mathcal{G}, n, \gamma\rangle$, we reformulate~\eqref{td_error} by introducing the aggregated reward, such that $r_{\textrm{team}}$ is the weighted sum of agents' reward, defined as follows:
\begin{align}
\label{reward}
r_{\textrm{team}} = \sum_{i\in\mathcal{V}}^{} \sum_{j\in\mathcal{E}_i}^{} \omega_{ij}r_j,
\end{align}
where $\mathcal{E}_i$ denotes the set of vertex indices that have an edge directed from $v_i$, and $r_j$ is the reward of the agent represented by $v_j$. 

In VDN, the main goal of each agent $i$ is to learn a policy that maximizes the overall team performance. Then, it is utilized to determine the best action for agent $a_i$ for a specific state $s$. This can be expressed as follows:
$$a_i = \pi^{\textrm{VDN}}_i(s) = \max_{a}(Q_i(s, a, w_i)), $$
where the learned policy of agent $i$ is denoted as $\pi_i$, and $Q_i$ represents its action-value function. This notation is applied to all agents in the following form:
$$u = \pi^{\textrm{VDN}}(s) = \max_{u}(Q(s, u, w)), $$
where $\pi$ represents a collection of policies, and $u$ is the joint actions of the agents. The formulation of G-VDN can be stated as follows:
$$u = \pi^{\textrm{G-VDN}}(s|\mathcal{G}) = \max_{u}(Q(s, u, w|\mathcal{G})), $$
where the policy is conditioned by the relational network $\mathcal{G}$. 

In cases where $\mathcal{G}$ represents a self-interest relational network, it is worth mentioning that the policies for G-VDN and VDN  become equivalent. This equivalence arises because, in VDN, each agent's reward is contributed equally to the overall team reward.
$$\pi^{\textrm{G-VDN}}(s|\mathcal{G}) \equiv \pi^{\textrm{VDN}}(s),$$
where $\mathcal{G} \in \{ \textrm{a}, \textrm{d}, \textrm{h} \}$ networks depicted in Fig.~\ref{fig:agent_relations}. Overall, G-VDN allows for the shaping and influencing of cooperation strategies among agents, whereas VDN leads to agents randomly converging to one of multiple solutions without any control over the team behavior.

\subsection{Training Phase}
\label{training_phase}
VDN employs a Deep Q-Network (DQN)~\cite{mnih2015human} to represent each agent's Q-function, enabling it to handle high dimensional state and action spaces in multi-agent systems. Yet, training a DQN can be challenging due to instability and divergence caused by updating the Q-network parameters in each step, violating the independently and identically distributed (i.i.d) data points assumption. To overcome these challenges, Mnih et al.~\cite{mnih2015human} also introduce several techniques such as experience replay and fixed Q-target networks, which have now become standard in many deep reinforcement learning algorithms.


Similar to VDN, we (i) maintain a replay memory to store the experiences, and randomly replay these experiences $m$ times during each training step, (ii) utilize two DQNs for each agent: a prediction network, and a fixed target network
which is the prediction network from a previous iteration. 
The training process of the networks entails three main components: generating episodes, updating the target network, and training the prediction network.

\noindent \textbf{Generating Episode:} An episode is generated by using the prediction network of each agent with $\varepsilon$-greedy approach. A large number of transitions experienced by each agent during the episode are stored in the replay memory. Each transition is a tuple $\langle S, A, R, S'\rangle$, where $S$ is a set of current states of the agents $\{s_1, s_2, \dots,  s_n\}$, $A$ contains the actions taken by each agent $\{a_1, a_2, \dots, a_n\}$, $R$ are the rewards received by the agents $\{r_1, r_2, \dots, r_n\}$ and $S'$ keeps the next state of the agents $\{s'_1, s'_2, \dots, s'_n\}$. The size of the memory is defined as a hyperparameter, and when it becomes full, the oldest transitions are replaced with new incoming ones.

\noindent \textbf{Target Network Update:} 
We use fixed target networks to increase the stability of agents' prediction network and prevent them from diverging. After a specified number of episodes have been generated, the weights of each agent's target network are updated with those of the corresponding agent's prediction network.
The idea of incorporating target networks is akin to having a calculated plan for achieving success in a game, and persisting with it until a superior one emerges, rather than altering one's strategy after each move. By allowing the network to consider 
multiple moves instead of continuously adjusting its parameters, it has the potential to develop a stronger model before being utilized for decision-making~\cite{mnih2015human}.  

\noindent \textbf{Prediction Network Training:} The prediction networks are trained each time an episode is generated. To train the prediction networks, a batch of transitions of size $b$ is sampled from memory. TD error is calculated using \eqref{td_error} for each instance in $b$ and the team reward used in the TD error is computed by utilizing \eqref{reward}. Lastly, the model parameters are updated using an optimizer in the direction that reduces the TD error. This step that includes sampling batches from the memory and updating the networks is repeated $m$ times per episode, where $m$ is a hyperparameter.

\section{Experiments}

\subsection{Environment}

To evaluate the effectiveness of the proposed approach in influencing and steering agents' behaviors, we applied G-VDN to the \textit{Switch} environment~\cite{magym} in three scenarios with varying numbers of agents both in simulation and real-world. The accompanying video shows the execution of the experiments in real-world\footnote{Accompanying video: \url{https://youtu.be/8m-rSsdloXU}}. As shown in  Fig.~\ref{fig:2_agent_results}(a), the \textit{Switch} environment consists of a grid-world that features two areas connected through a narrow passageway. The \textit{Switch} environment in its original size accommodates a maximum of four agents marked with different colors that are trying to reach their respective goals, marked in boxes of the same color as each agent. Each agent's goal cell is positioned on the opposite side of its initial location, posing a challenge given that only one agent can pass through the passageway at a time. Agents can take one of the five actions for each time step: move left, right, up, down, or stay still. Upon reaching their goal, they are awarded $+5$ points. The agents are penalized with $-0.1$ points for each time step spent away from their goal. The episode concludes once all agents have arrived at their respective goal locations or after a maximum of $50$ time steps. To successfully reach their goal locations, the agents must cooperate to avoid blocking each other's paths while maximizing total team rewards.
We chose the \textit{Switch} environment because firstly, learning an optimal behavior requires cooperation among the agents. Secondly, the presence of the bridge allows for easily observable and measurable agents' behaviors and facilitates the evaluation of relational networks effects in different scenarios. And finally, since \textit{Switch} was used in the original VDN paper, we can provide a clear comparison between our method and VDN. 

\subsection{Models and Hyperparameters}
In our setup for the \textit{Switch} environment, we configured our algorithm as follows: We used a Multi-Layer Perceptron (MLP) with two hidden layers, each consisting of $128$ neurons and utilizing the ReLU activation function. For training each agent's prediction model, we performed $m=10$ iterations per episode using batches of size $b=32$, randomly sampled from a replay memory with a capacity of $50$k time-steps. We employed the \textit{Adam} optimizer with a learning rate of $0.001$ and utilized the squared TD-error as the loss function. The weights of the target network were updated every $200$ episodes with the weights of the prediction network. To encourage exploration, we employed the  $\varepsilon$-greedy method, where $\varepsilon$ linearly decreased over time. Finally, we set the discount factor ($\gamma$) to 0.99.



\begin{figure}[t]
\centering
  \includegraphics[width=\linewidth]{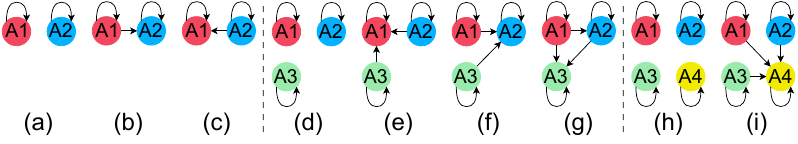}
  \caption{\small{Relational networks used in G-VDN. 
  The edges represent whether an agent puts importance on another agent, including themselves. The relational networks in (b-c) were used in \textit{two-agent} experiments in section~\ref{two-agents}; (e-g) were used in \textit{three-agent} experiments in section~\ref{three-agents}; (i) was used in the \textit{four-agent} experiment in section~\ref{four-agents}.
  }
  } 
  \label{fig:agent_relations}
\end{figure}

\begin{figure*}[t]
\centering
  \includegraphics[width=\linewidth]{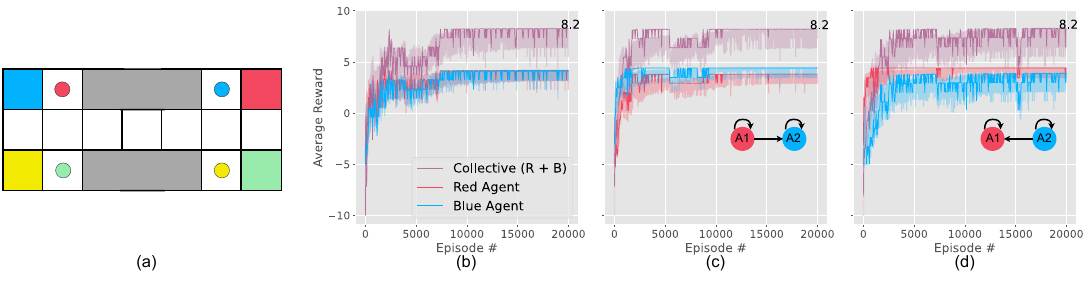}
  \caption{\small{(a) \textit{Switch} environment with four agents. In the three-agent setup, the yellow agent and its goal location were removed. For the two-agent setup, the green agent and its goal location were in addition removed. Two-agent experiment results: (b) VDN, (c-d) G-VDN with relational networks in Fig.~\ref{fig:agent_relations}(b) and Fig.~\ref{fig:agent_relations}(c), respectively.}} 
  \label{fig:2_agent_results}
\end{figure*}

\subsection{Results}

As the environment can accommodate up to four agents, we have designed three different setups for our experiments, comprising of two, three, or four agents. This has resulted in six independent experiments. In all experiments, we compared our algorithm against VDN over the individual and collective rewards collected by the team.


In all the figures that depict the results of the experiments (i.e., Figs.~\ref{fig:2_agent_results}, \ref{fig:3_agent_results}, and  \ref{fig:4_agent_results}), the shaded regions display the average training reward over 10 runs, while the solid lines represent the average test rewards of the agents. The test rewards are determined by interrupting the training every 50 episodes and evaluating the individual rewards of the agents using the greedy strategy. In Figs.~\ref{fig:2_agent_results}, \ref{fig:3_agent_results}, and  \ref{fig:4_agent_results}, collective reward depicts the sum of individual rewards that is achieved by the agents in the environment. Note that collective reward is simply the addition of the individual agents' rewards (i.e., the team reward maximized in VDN), which is different from the team reward which is used for training G-VDN.


\subsubsection{Two-Agent Scenario} \label{two-agents}
This scenario consists of two agents represented by the red and blue circles and their respective goal locations marked with the same colors as shown in Fig.~\ref{fig:2_agent_results}(a). To successfully reach their respective goal locations, one agent must learn to wait until the other agent passes the bridge. Therefore, there are two optimal policies for the agents to follow: either the red agent passes the bridge first or the blue agent does. In each run, VDN converges to one of the two optimal policies that the algorithm has experienced first based on the randomness in the exploration phase. 
As seen in Fig.~\ref{fig:2_agent_results}(b) and Table~\ref{experiment_results}, the individual rewards for both agents are nearly identical as a result of averaging their rewards over 10 runs. In other words, in some runs, the red agent learns to cross the bridge first, leading to the maximization of its individual reward, while in others, the blue agent crosses the bridge and maximizes its own reward. Next, we designed two relational networks for G-VDN to better understand the influence of team interactions on optimal solutions. Specifically, in one relational network, the red agent assigns importance to the blue agent (Fig.~\ref{fig:agent_relations}(b)), while in the other relational network, the blue agent places importance on the red agent (Fig.~\ref{fig:agent_relations}(c)). The directed edges indicate whether an agent places importance on another agent, which dictate whether an agent yields to the other agent and lets them pass the bridge first for this environment. 
The results in Fig.~\ref{fig:2_agent_results}(c) and Table~\ref{experiment_results} show that when training the network of agents with G-VDN and using the relationships shown in Fig.~\ref{fig:agent_relations}(b), the blue agent accumulates higher individual rewards ($4.40\pm0.00$) as it is the first to use the bridge to reach its goal than the red agent ($3.80\pm0.00$). Conversely, when Fig.~\ref{fig:agent_relations}(c) was applied, the situation is reversed with the red agent having a higher total reward ($4.39\pm0.02$), as depicted in Fig.~\ref{fig:2_agent_results}(d). Comparing the collective rewards from these three experiments reveals that they are all the same as the graph steers which \textit{optimal} solution the algorithm converges to.

\begin{table*}[t]
\centering
\caption{
Average reward with 95\% confidence intervals for ten runs after training completed.
}
\resizebox{\textwidth}{!}{
\begin{tabular}{cccc|cccc|cc}
\hline
                                       & \multicolumn{3}{c|}{Two-Agent Scenario}  & \multicolumn{4}{c|}{Three-Agent Scenario}     & \multicolumn{2}{c}{Four-Agent Scenario} \\ \cline{2-10} 
                                       & VDN         & \multicolumn{2}{c|}{G-VDN} & VDN       & \multicolumn{3}{c|}{G-VDN}        & VDN                & G-VDN              \\ \cline{2-10} 
\multicolumn{1}{l}{Relational Network} & N/A         & Fig.~\ref{fig:agent_relations}(b)          & Fig.~\ref{fig:agent_relations}(c)         & N/A       & Fig.~\ref{fig:agent_relations}(e)       & Fig.~\ref{fig:agent_relations}(f)       & Fig.~\ref{fig:agent_relations}(g)       & N/A                & Fig.~\ref{fig:agent_relations}(i)                \\
Red Agent                                    & 4.10±0.19   & 3.80±0.00    & 4.39±0.02   & 4.38±0.04 & 4.40±0.00 & 3.75±0.03 & 3.13±0.03 & 3.70±0.34          & 3.43±0.29          \\
Blue Agent                                  & 4.15±0.15   & 4.40±0.00    & 3.86±0.06   & 3.66±0.04 & 3.70±0.00 & 4.40±0.00 & 3.78±0.02 & 3.78±0.25          & 3.61±0.17          \\
Green Agent                                 & ---         & ---          & ---         & 4.35±0.02 & 4.30±0.00 & 3.76±0.07 & 4.40±0.00 & 3.77±0.36          & 3.54±0.23          \\
Yellow Agent                                & ---         & ---          & ---         & ---       & ---       & ---       & ---       & 3.80±0.30          & 4.34±0.09          \\ \hline
\end{tabular}
}
\label{experiment_results}
\end{table*}

\subsubsection{Three-Agent Scenario} \label{three-agents}

\begin{figure}[!b]
\centering
  \includegraphics[width=\linewidth]{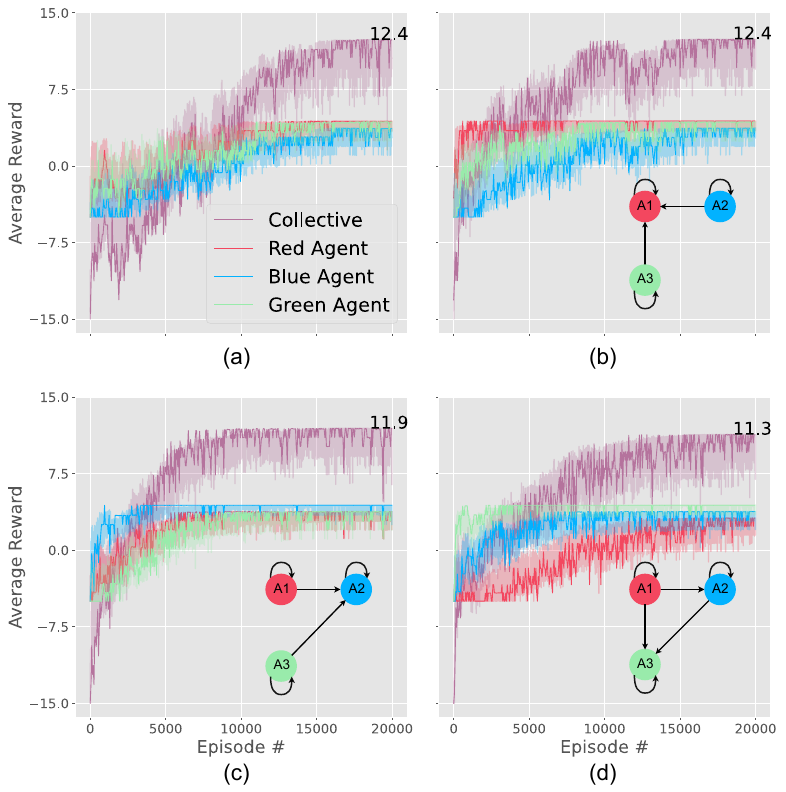}
  \caption{\small{Three-agent experiment results. (a) VDN, (b-d) G-VDN with relational networks in Fig.~\ref{fig:agent_relations}(e), Fig.~\ref{fig:agent_relations}(f), and Fig.~\ref{fig:agent_relations}(g), respectively.
  }}
  \label{fig:3_agent_results}
\end{figure}

In this scenario, the \textit{Switch} environment used with three agents: the red and green agents are on the left and the blue agent is on the right side of the bridge as shown in Fig.~\ref{fig:2_agent_results}(a). The optimal policy, minimizing waiting time, is for the pair of agents situated on the left to cross the bridge initially, as they are heading in the same direction and can proceed consecutively. Yet, because either agent can go first (red or green), this results in two optimal policies for crossing the bridge, either red--green--blue or green--red--blue. Fig.~\ref{fig:3_agent_results}(a) and Table~\ref{experiment_results} depict the results obtained with VDN, where such preferences of order cannot be specified. The results show that the average individual rewards of the red ($4.38\pm0.04$) and green ($4.35\pm0.02$) agents are similar to each other, as they randomly alternate taking the lead in crossing the bridge for the $10$ runs, and reach higher scores than those of the blue agent ($3.66\pm0.04$). However, G-VDN can influence the order using the relational network depicted in Fig.~\ref{fig:agent_relations}(e), where the red agent has the highest importance. As demonstrated in Fig.~\ref{fig:3_agent_results}(b) and Table~\ref{experiment_results}, the order of bridge usage becomes consistent as red--green--blue with the individual rewards of ($4.40\pm0.00$), ($4.30\pm0.00$), and ($3.70\pm0.00$), respectively.

We have demonstrated that the type of \textit{optimal} solution the algorithm converges to can be guided via relational networks. G-VDN, in addition, enables the agents to discover new behaviors to converge to based on the inter-agent relationship. For instance, the relationships in Fig.~\ref{fig:agent_relations}(f) show that the red and green agents prioritize the blue agent, granting it the first turn. The second turn is then randomly assigned between the red and green, as their importance weights are equal. The results of training the agents with G-VDN using this relational graph are shown in Fig.~\ref{fig:3_agent_results}(c), which indicates that the blue agent's ($4.40\pm0.00$) individual reward is higher than the others' and the red ($3.75\pm0.03$) and green ($3.76\pm0.07$) agents' rewards are almost the same due to alternating turns. 
To examine the extent to which the agents follow the relational network and prioritize defined cooperation strategy instead of solely maximizing the collective and individual rewards, we define a new relational graph shown in  Fig.~\ref{fig:agent_relations}(g). The graph elicits the following sequence of crossing the bridge: green(left)--blue(right)--red(left), causing the red agent to wait the alternating turn instead of crossing the bridge a step after the green agent. Figure~\ref{fig:3_agent_results}(d) and Table~\ref{experiment_results} show that the agents followed the defined relationships, demonstrating their ability to prioritize relationships over collective reward, even at a cost. It should be noted that, among the experiments in this section, the lowest collective reward is obtained with this relational graph as it requires the red agent to sacrifice more of its reward by taking the final turn.

\begin{figure}[b]
\centering
  \includegraphics[width=\linewidth]{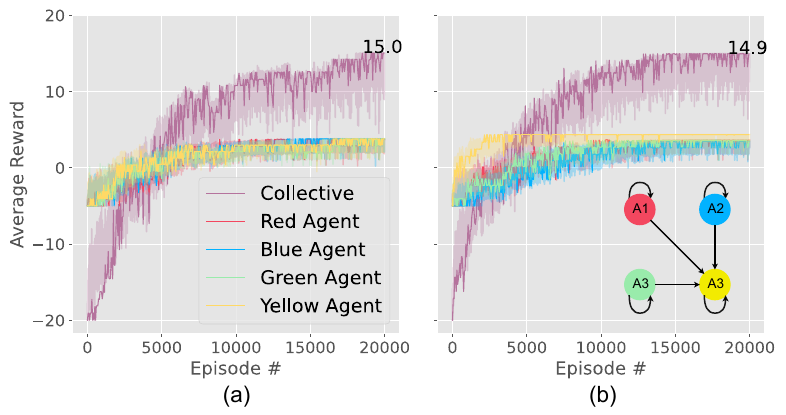}
  \caption{\small{Four-agent experiment results. (a) VDN, (b) G-VDN with relational network in Fig.~\ref{fig:agent_relations}(i).}}
  \label{fig:4_agent_results}
\end{figure}

\subsubsection{Four-Agent Scenario} \label{four-agents}
The final experiment involves four agents in the \textit{Switch} environment as shown in Fig.~\ref{fig:2_agent_results}(a) with equal numbers of agents each side. Applying VDN does not guarantee a deterministic solution for which side crosses the bridge first or which agent on the side takes the first turn. Instead, it leads to a stochastic selection process for these decisions, resulting in the learned team behavior varying from one run to the next.
Our results depicted in Fig.~\ref{fig:4_agent_results}(a) confirm that according to VDN, the individual rewards are very close (similar) due to the randomness caused by having multiple solutions that equally place the importance on all the agents without specifying any relationships (See Table~\ref{experiment_results}).

Yet, when using G-VDN, the relational network shown in Fig.~\ref{fig:agent_relations}(i) influences the order in which the bridge should be navigated. Since all agents place the importance to the yellow agent, the right side learns to take the first turn and the yellow agent is the first to use the bridge. This behavior is illustrated in Fig.~\ref{fig:4_agent_results}(b). Similar to the three-agent scenario, we expect that the blue agent crosses the bridge after the yellow agent to compensate for the negative reward of remaining in the environment. However, due to the increased complexity of the team network and small amount of negative reward, this behavior has been observed in 80\% of the runs (8/10 runs). In the other 20\%, the bridge was crossed by either of the three agents randomly. As shown in Fig.~\ref{fig:4_agent_results}(b) and Table~\ref{experiment_results}, this behavior causes the individual rewards for the blue, red, and green agent to be very similar, while the yellow agent to attain the highest reward ($4.34\pm0.09$). Between the green and the red agent, the order of using the bridge was determined randomly (since they have the same weight), resulting in similar individual rewards. 

\subsection{Real-World Experiments}

To validate the effectiveness of our proposed algorithm in a real-world environment, we replicated the \textit{Switch} grid-world in the lab as illustrated in Fig.~\ref{fig:robots_in_action}. We employed four Turtlebot4 mobile robots marked by different colors, identical to our simulation experiments. Robots communicate through the Robotic Operating System (ROS) and a separate identification parameters has been assigned to each robot. We used a graph-based SLAM algorithm~\cite{macenski2021slam} to map the environment and localize the robots. Based on this map, we determined the coordinate of the cells in the \textit{Switch} environment. In each time-step, the robots receive commands generated by (the trained) G-VDN through ROS to either stay in their current cell or move to a neighboring cell. The accompanying video shows the execution of the tasks.

\section{Conclusion and Future Work}

We have proposed a novel algorithm that utilizes a relational network to influence agents' behavior within a team, leading to new cooperative behaviors. Given a team structure, our algorithm discovers and guides cooperation strategies that are driven by agents' relationships. As our results demonstrate, our algorithm is effective in influencing team behavior according to the specified relational network even in real-world multi-robot applications that require team cooperation.



Our proposed approach also has a few inherent limitations that open up novel research challenges for future work. Although in this paper, we consider the team structure to be given, methods for discovering the team structure could be investigated. For instance a high-level cost (e.g., observed behavior and capability differences of agents) can be used to find an optimal structure. Another challenging aspect of the proposed approach is that the relational network becomes denser as the number of agents increases. Thus, it becomes essential to develop methods for dealing with the tedious and eventually impractical task of adjusting relational weights. 


\section*{Acknowledgement}
This work is supported in part by NSF (IIS-2112633) and
the Army Research Lab (W911NF20-2-0089).

\addtolength{\textheight}{-8.1cm} 

\bibliographystyle{IEEEtran}
\bibliography{IEEEabrv, refs}

\end{document}